\documentclass[twocolumn,10pt]{asme2e}

\usepackage{graphicx}      
\usepackage{siunitx}
\usepackage{enumerate}

\usepackage{tikz}
\usepackage{float}
\usepackage[utf8]{inputenc}
\usepackage{algorithm}
\usepackage[noend]{algpseudocode}
\usepackage{comment}
\usepackage{epstopdf}
\usepackage{comment}
\usepackage{subfig}
\usepackage{pgfplots}
\usepackage{amsmath,amsfonts,amssymb}
\usepackage{amsmath}             
  {

  }
\usepackage{graphicx}
\usepackage{caption}
\usepackage{threeparttable}
\makeatletter
\newcommand*\bigcdot{\mathpalette\bigcdot@{.5}}
\newcommand*\bigcdot@[2]{\mathbin{\vcenter{\hbox{\scalebox{#2}{$\m@th#1\bullet$}}}}}
\makeatother

\conffullname{the 2022 International Symposium on Flexible Automation\\ ISFA2022}

\confdate{July 3-7}
\confyear{2022}
\confcity{Yokohama}
\confcountry{Japan}

\papernum{ISFA2022-C000082}

\title{Robust Task Planning for Assembly Lines with Human-Robot Collaboration}

\author{ Jessica Leu\thanks{Contact author: \tt\small jess.leu24@berkeley.edu.} , Yujiao Cheng, Masayoshi Tomizuka
    \affiliation{
	Department of Mechanical Engineering\\
	University of California\\
    Berkeley, CA 94720 USA}
    
}

\author{ Changliu Liu
    \affiliation{
	Robotics Institute\\
	Carnegie Mellon University\\
    Pittsburgh, PA 15213 USA}	
}

\begin{document}

\maketitle    

\begin{abstract}
{\it Efficient and robust task planning for a human-robot collaboration (HRC) system remains challenging. The human-aware task planner needs to assign jobs to both robots and human workers so that they can work collaboratively to achieve better time efficiency. However, the complexity of the tasks and the stochastic nature of the human collaborators bring challenges to such task planning. To reduce the complexity of the planning problem, we utilize the hierarchical task model which explicitly captures the sequential and parallel relationships of the task. To account for human-induced uncertainties, we model human movements with the sigma-lognormal functions. A human action model adaptation scheme is applied during run-time and it provides a measure for modeling the human-induced uncertainties. We propose a sampling-based method to estimate the uncertainties in human job completion time. Next, we propose a robust task planner, which formulates the planning problem as a robust optimization problem by considering the task structure and the uncertainties. We conduct simulations of a robot arm collaborating with a human worker in an electronics assembly setting. The results show that our proposed planner, compared to the baseline planner, can reduce task completion time when human-induced uncertainties occur.}
\end{abstract}


\section{INTRODUCTION}

The development of intelligent industrial robots is craving for an efficient, reactive, and robust task planning in dynamic environments. For example, task planners for human-robot collaboration (HRC) systems in assembly lines \cite{nikolaidis2012human, coupete2015gesture} need to decide in real time how to assign different jobs to both human workers and robots to minimize task completion time. These HRC applications, however, raise three challenges for robotic systems: 1) predicting humans' actions and intentions \cite{cheng2019human, leu2019motion, cheng2020towards}, 2) taking humans' actions into account in the planning problem \cite{cheng2021human}, and 3) designing a computationally efficient planner.

In HRC environments, it is important to predict the duration of the human actions to make a collaborative task plan. We adopt our previous work \cite{cheng2021long} and use the sigma-lognormal function to model the human movement. An online adaptation process adapts the human action model according to the new observations. This method enables us to predict the human trajectory and estimate the duration for the human's current action and future actions. More importantly, we reckon that the adaptation process provides us information to estimate the uncertainty of the duration prediction. Intuitively, the more the human action model needs to be adapted, the less we can trust the duration prediction before the adaptation converges to a new human action model. We propose a sampling process to estimate the human action duration uncertainty.   

The goal of the human-aware task planning is to assign the robot an action from all feasible actions to minimize the collaborative costs including factors such as completion time \cite{gombolay2018fast}, human fatigue \cite{li2019sequence}, and spatial interfaces \cite{gombolay2018fast}. In order to plan efficiently, these planners require some task knowledge for constructing the task model. The two popular task models include flat models, such as a plan network \cite{levine2014concurrent}, or hierarchical models, such as and/or graphs \cite{knepper2014distributed, boyd2019hierarchical}. Hierarchical models have shown superiority over flat models when used for predicting human actions and planning predictable actions for the robot. 
In this work, we adopt our previously proposed sequential/parallel task model to facilitate task-level prediction and optimization-based planning \cite{cheng2021human}. This framework also allows us to incorporate human-induced uncertainties into the planning problem. 

Previous works on HRC task planning mainly address uncertainties for safety considerations \cite{johannsmeier2016hierarchical,mainprice2013human}. However, one should not overlook the importance of accounting human motion uncertainties for time efficiency of the task plan. With the task model and the duration uncertainty estimation, we proposed a robust optimization-based formulation, in which the objective is to minimize the completion time of the planning horizon. This problem can be formulated as a mixed-integer linear programming problem, which can be solved efficiently. Simulations of a computer assembly task with different human behaviors are conducted to verify the effectiveness of the proposed robust task planner, and the performance is compared with that of the baseline planner.
The key contributions of this work are: 
\begin{enumerate}
    \item We propose a sampling-based estimator that utilizes information from the human action model adaptation process to estimate human action duration uncertainty.
 	\item We formulate the robust task planning problem as a robust optimization problem.
 	\item Simulations are conducted to verify the performance of the robust task planner.
\end{enumerate}

\section{RELATED WORKS}
The goal of the online task planner considered in this work is to allocate actions to both robots and human workers. Many previous works address these problems by constructing either a tree \cite{cheng2021human,cirillo2009human} or a graph \cite{levine2014concurrent,johannsmeier2016hierarchical} and use search-based methods to find the best plan according to the given objectives. Human observations are often used to prune unpromising edges in the tree or graph for better planning efficiency and quality \cite{cirillo2009human,levine2014concurrent,giele2015dynamic}. Some other works formulate the task planning problem as a multi-agent planning problem and apply reinforcement learning to learn a cooperative policy \cite{koppula2016anticipatory,ghadirzadeh2020human}. While human motion is taken into account for safety and ensuring plan validation, to the best of the authors knowledge, previous works have not explicitly considered the effect of uncertainty in human action duration and effect on task completion time. Previously, most works treat the estimated human action duration, or human action completion time, of each individual action as a given constant \cite{cirillo2009human,giele2015dynamic,johannsmeier2016hierarchical}. However, the original plan based on the estimated time information may no longer be time-optimal due to the varying human action completion time during execution. Since the objective of most task planners is to minimize task completion time, we claim that it is important to address the uncertainty in human action duration in the planning, which is the main focus of this work. 
\section{PRELIMINARIES}
\subsection{Hierarchical Task Model}
\begin{figure}[t]
	\begin{center}
		\includegraphics[width=.9\linewidth]{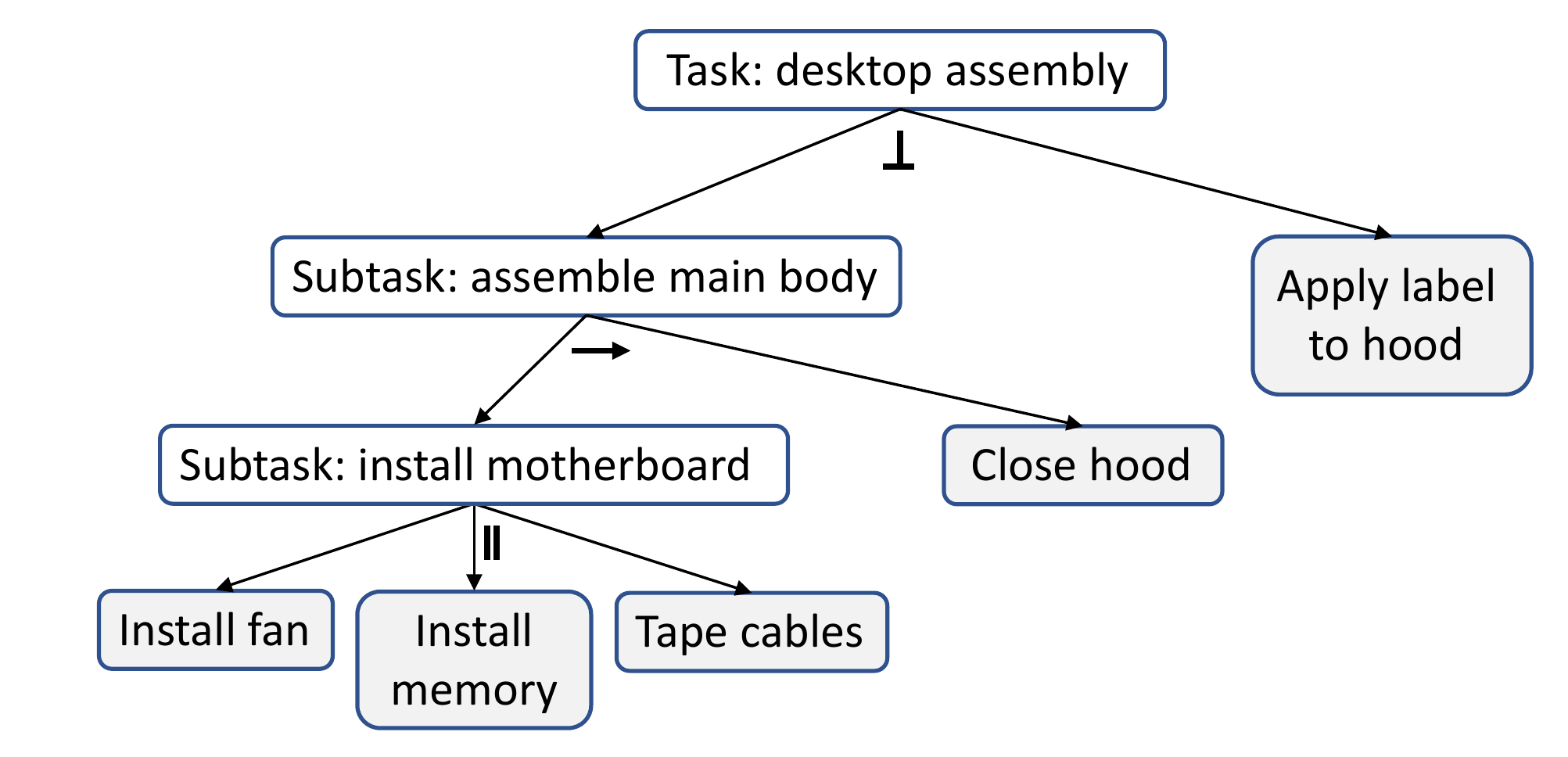}
		\caption{The sequential/parallel task model for a desktop assembly task.}
		\label{fig: example}
	\end{center}
\end{figure}

In this work, we target the computer assembly task \cite{cheng2021human}. Fig.~\ref{fig: example} shows an example of the sequential/parallel task model for the target task. The root node represents the \textit{task}, the leaf nodes (colored in gray) are \textit{actions}, and all the other nodes represents \textit{subtasks}. The indicators below the root and the subtasks nodes indicates the relationship among their child nodes. The three types of relationships are:\\
\textbf{Sequential nodes}: their child nodes must be executed in the order from left to right, which is denoted by the operator $\to$. For example, the subtask \emph{assemble main body} is a sequential node, and its child \emph{install motherboard} must be done before \emph{close hood}.\\
\textbf{Parallel nodes}: their child nodes can be executed in parallel, which is denoted by $\parallel$. For example, the subtask \emph{install motherboard} is a parallel node, its children \emph{install CPU fan}, \emph{install memory}, and \emph{install memory} can be executed simultaneously. \\
\textbf{Independent nodes}: their child nodes can be executed in any orders, which is denoted by $\perp$. Parallel nodes are special case of independent nodes. For example, root node is an independent node but not a parallel node. Its child nodes \emph{applying labels to hood} and \emph{assemble main body} have no fixed order, but they cannot be executed in parallel if \emph{close hood} is in progress.

Following the description of \cite{cheng2021human}, actions can be defined as $a = [\{motion, object\}, attribute]$, where \textit{motion} indicates the types of the movement, \textit{object} indicates the object of interaction, and \textit{attribute} contains information such as completion time and energy consumption, which are useful in the planning process.

\subsection{Sigma-lognormal Model}
Sigma–lognormal model can explain most of the basic phenomena on human motor control \cite{plamondon2003kinematic} and can be used to model human motions in the assembly setting \cite{cheng2021long}. The human action model takes the following form:
\begin{equation}\label{eq:log-normal}
\begin{aligned}
    &\hat{\vec{v}} (t)   = \sum_{i=1}^N \hat{\vec{v}}_i (t) =  \sum_{i=1}^N \vec{D}_i (t) \Lambda_i(t; t_{0i},\mu_i, \sigma^2_i),\\
    &\Lambda_i(t; t_{0i},\mu_i, \sigma^2_i) = \frac{1}{\sigma_i \sqrt{2\pi}(t-t_{0i})}exp( \frac{-(\ln(t-t_{0i})-\mu_i)^2}{2\sigma^2}),
\end{aligned}
\end{equation}
where $\hat{\vec{v}}(t)$ is the velocity of the human hand at time $t$ and $\Lambda(t, t_{0},\mu, \sigma^2)$ is a lognormal distribution with the time shift $t_{0}$, the expected value of the $t$'s natural logarithm $\mu$ and the standard deviation of the $t$'s natural logarithm $\sigma$. The velocity $\hat{\vec{v}}(t)$ is composed of $N$ lognormal distributions $\Lambda_i$, each scaled by variable $\vec{D}_i, i = 1,...,N$. Each action in the task can be modeled by its own sigma-lognormal human action model. In this work, we use $N=2$ to capture motions in 2D space, i.e., $i\in{x,y}$. 

To obtain the parameters $\vec{\alpha}_i = \{\vec{D}_i, t_{0i}, \mu_i, \sigma_i^2\}, i = 1,...,N$, of the nominal human action model, we collect training data of similar action motions and use the Levenberg–Marquardt algorithm \cite{seber2003nonlinear} to solve the following problem: given a set of $m$ data points $(t_{p,j}, {\vec{v}}_{p,j})$ from the $p$-th collected trajectory, find $\vec\alpha$ such that the sum of the squares of the deviations for all $q$ collected trajectories  $S(\vec\alpha)$ is minimized:
\begin{equation}\label{eq:fit-model}
\begin{aligned}
{\vec\alpha^*} = \arg\min_
{\vec\alpha} S(\vec\alpha) = \arg\min_{\vec\alpha} \sum_{p=1}^q \sum_{j=1}^m ||{\vec{v}}_j - \hat{\vec{v}}(t_j, \vec\alpha)||_2^2.
\end{aligned}
\end{equation}

\subsection{Human Action Model Adaptation}\label{sec:adp}
It is necessary to adapt the offline learned human action models to accommodate different human motion styles during action executions. \cite{woch2011kinematic} proposed to adapt the human action model by time scaling and shifting, i.e., modifying $\mu_i$ and $t_{0i}$ in~\eqref{eq:log-normal} through scaling factor $S_{t,i}$ and $s_{t,i}$. In addition, \cite{cheng2021long} proposed to scale $D_{i}$ with $S_{D,i}$ to account for more variations.
\begin{equation}\label{eq:adp}
\begin{aligned}
    &t_{0s,i} = S_{t,i} t_{0i} + s_{t,i},\\
    &\mu_{s,i} = \mu_{i} + ln(S_{t,i}), \\
    &D_{s,i} = S_{D,i}D_{i}.
\end{aligned}
\end{equation}
Therefore, the sigma-lognormal human action model becomes $\hat{\vec{v}}(t; \vec{\alpha}^{*}, \vec{\beta})$, where $\vec{\beta}_i = \{S_{t,i}, s_{t,i}, S_{D,i}\}, i = 1,...,N$. To adapt $\vec{\beta}$ when a new data point is available, the human action model is first updated by using Levenberg–Marquardt algorithm that minimizes the prediction error, which gives the update rule: 
\begin{equation}
\vec{\beta}_i=\vec{\beta}_i - (\hat{v}_{i, {t_k}} - {{v}}_{i, {t_k}})\nabla_{\vec{\beta}_i}\hat{v}_{i, {t_k}}./((\nabla_{\vec{\beta}_i}\hat{v}_{i, {t_k}})^{.2} + \vec{\lambda}_i),\: i=1,\dots,N,\label{eq:grad1}\\
\end{equation}
where $v_{i, {t_k}}$ is the measurement and $\hat v_{i, {t_k}}$ is the model predictions on either the $x$ or the $y$ direction at time $t_k$. $\nabla_{\vec{\beta}_i}$ is the gradients of $\hat{v}_{x, {t_k}}$ with respect to $\vec{\beta}_i$, and $\vec{\lambda}_i$ is the non-negative damping factors. Note that the operator ``.'' indicates the element-wise operation. The zero-crossing time after this model update process will be used as the estimated final time $\hat t_f$, also called the human action duration or the human action completion time. Second, to take advantage of the scene information, $\vec{\beta}$ is updated to minimize an objective function with three terms: 1) the difference between the current distance to the goal and the predicted travel distance to the goal; 2) prediction error at the current time $t_k$; 3) velocity at the final time $\hat t_f$. Therefore, to optimize the objective functions, the two ends of the future velocity profile are fixed and the velocities are modified in between to best fit the scene information.
The objective function is a weighted sum of the three terms: $K(\hat t_f, \vec{{\beta}}) = \gamma_1 J_1(\hat t_f,\vec{\beta}) +\gamma_2 J_2(\vec{\beta})+\gamma_3 J_3(\hat t_f,\vec{\beta}) $,
and the update rule is:
\begin{equation}
\vec{\beta}_i= \vec{\beta}_i+K(\hat t_f, \vec{\beta})\nabla_{\beta_i}K./((\nabla_{\beta_i}K)^{.2}+\vec{\lambda}'_i),\: i=1,\dots,N.\label{eq:grad2} \\
\end{equation}
The parameter $\vec{{\beta}}$ can be view as the characteristics of the human worker such as the worker's tendency to execute every action quickly or slowly. Assuming that the worker carries the same characteristics when executing all the actions, we can directly apply $\vec{{\beta}}$ from one model to scale the other models for potentially more accurate prediction of those action completion time, as suggested in \cite{cheng2021long}. 

In this work, we proposed a sampling based method that utilizes the gradients $\nabla_{\vec{\beta}_i}\hat{v}_{i, {t_k}}$ and $\nabla_{\vec{\beta}_i}K$ to estimate the human-induced uncertainty.

\section{METHODS}
\subsection{Baseline Problem Formulation}
In this work, the human-aware robot task planning problem considers action allocation problems for a robot and a human worker. Notice that the proposed formulation and methods can be easily extended to the multi-robot case. The number of actions considered in one planning problem is determined by the number of the remaining actions that are parallel to the human's current action. Thus, the planning horizon $k$ is part of the remaining actions and it varies as the task proceeds. During run-time, the planner identifies the human's current action, then find the parallel actions according to the hierarchical task model. The planning problem can then be formulated for assigning those parallel actions to the human and the robot such that the completion time for the planning horizon $t$ is minimized. Denote the $k$-dimensional action assignment vectors $x_r,x_h\in \{1,0\}^k$ for the robot and the human, respectively, where the entries are either $0$ (not assigned) or $1$ (assigned). For example, $x_r = [1,0]^\top$ and $x_h = [0,1]^\top$ together means that there are two actions in the planning horizon. The first action is assigned to the robot while the second action is assigned to the human worker. The inputs of the algorithm are $T,C,C_r,C_h,t_0,t_r$, and $t_h$, where $T$ is the sequential/parallel task model, $C = \{1,\dots,k\}$ is the set of action indices, $C_h$ and $C_r$ are the actions indices that human worker and the robot are capable of executing, respectively, $t_o\in \mathbb Z_{+}$ is the remaining time before the human's current action is completed, and $t_h,t_r \in \mathbb Z_
{+}^k$ are the empirical completion time of a human and a robot for each action, respectively. If $t_r$, $t_0$, and $t_h$ are time invariant, the optimization problem is formulated as follows: 
\begin{equation}\label{eq:TI_opt}
\begin{aligned}
\min_{x_h,x_r,t} \quad & t, \\
\textrm{s.t.} \quad & x_h^\top t_h + t_o \leq t,  && x_r^\top t_r \leq t,   \\
  & x_h + x_r = \textbf{1}, && x_{h,i}, x_{r,i}\in\{0,1\},& \;i = 1,\dots,k, \\
  & x_{h,\{C-C_h\}}=\textbf{0}, && x_{r,\{C-C_r\}}=\textbf{0}.\\
\end{aligned}
\end{equation}
Decision variables $x_h$ and $x_r \in \{0,1\}^k$ are binary vectors for the assignment of actions, where $k$ is the number of actions in the planning horizon. 
Objective function $t$ is the completion time for the planning horizon, which is the upper bound of the human's completion time and the robot's completion time. $\{C-C_h\}$ and $\{C-C_r\}$ denotes the indices of the actions that the human and the robot is unable to do, respectively. During run-time, the planner optimizes the planning problem, then sends the selected action, which has the shortest completion time among all actions assigned to the robot, to the controller for execution, then plans again, repeatedly. 

It is clear that the optimality of the plan holds only when $t_0$ and $t_h$ are the same as the true action completion time. However, this is seldom the case. Even though the action completion time of a group of workers often stay in a range and one can use the longest time duration for solving all planning problem, the planner could be unnecessarily conservative while still cannot account for abnormal human behaviors, leading to a task completion time far longer than the time truly needed. Therefore, we propose to include such human uncertainty into the planning problem by extracting the human uncertainty information during the human action model adaptation process, then incorporate the completion time uncertainty into the the planning problem.    

\subsection{Task Planning with Human-induced Uncertainties}
Human-induced uncertainties affect HRC systems mainly from the safety aspect and the time efficiency aspect. While many works have addressed these uncertainties for safety \cite{cheng2019human,  leu2019motion, leu2020safe}, few has attempted to address the impacts on time efficiency caused by these uncertainties. To address this problem, we first reckon that human-induced uncertainty mainly affects $t_0$ and $t_h$. In this light, the task planning problem should be rewritten as: 
\begin{equation}\label{eq:TV_opt}
\begin{aligned}
\min_{x_h,x_r,t} \quad & t, \\
\textrm{s.t.} \quad & x_h^\top t_h + t_o \leq t,  && x_r^\top t_r \leq t,&   \\
  & x_h + x_r = \textbf{1}, && x_{h,i}, x_{r,i}\in\{0,1\},& \;i = 1,\dots,k, \\
  & x_{h,\{C-C_h\}}=\textbf{0}, && x_{r,\{C-C_r\}}=\textbf{0},& \\
  & t_0  = \bar t_0 + u_0, && \;u_0 \in\mathcal U_0,\\
  & t_h  = \bar t_h + u_1, && \;u_1 \in\mathcal U_1,
\end{aligned}
\end{equation}
where $\bar t_0$ and $\bar t_h$ are the originally proposed action completion time; $u_o$ and $u_1$ are added to account for the uncertainty on human action completion time; $\mathcal U_0$ and $\mathcal U_1$ are the uncertainty sets. To solve this optimization problem efficiently, a more solvable form of the problem is needed. Two questions are to be addressed: 
\begin{enumerate}
    \item To what extent should the plan be immune to the uncertainty?
 	\item What form should $\mathcal U_0$ and $\mathcal U_1$ take to model the action completion time uncertainties?
\end{enumerate}

To answer the first question, as mentioned in the literature \cite{ben2009robust}, it is sufficient to find a solution plan that is immune to all disturbances from $\mathcal U$ so that it is also immune to ``nearly'' all real-world disturbances, i.e., up to $(1-\epsilon)$ of the total probability mass, where $\epsilon$ is a small number. This can be achieved by finding a solution of a chance constraint problem, in which we need to choose $\mathcal U$ as a computationally tractable convex set that ``($1-\epsilon$)-supports'' all real-world disturbances. In other words, we answer the first question by formulating the first constraint as a chance constrain, i.e., $Prob(x_h^\top t_h + t_o \leq t)\geq (1-\epsilon)$ and choose a small $\epsilon =0.005$, so that we can say the solution plan guarantees that the completion time for the planning horizon will be smaller or equal to $t$ for $99.5 \%$ of the time. The second question can be answered by following the robust optimization formulation described in \cite{ben2009robust}, which requires an estimate of the human-induced uncertainty in order to construct the uncertainty set.     
\subsection{Human-induced Uncertainty Estimation}
In this section we proposed a sampling-based method to estimate the human-induced uncertainty. While human-induced uncertainty may be an abstract concept, we can focus on the ``correctness'' of the current human action model and use it to construct the uncertainty set. The basic idea behind the proposed method is that the less correct the action model is, the more adaptation it needs since it is deviating from the observation and the task scene, and therefore, the less we should trust the human action model and the originally proposed action completion time $\bar t_0$ and $\bar t_h$. In other words, larger uncertainty sets should be applied to $\bar t_0$ and $\bar t_h$ when the action model is adapting. Recall that in section~\ref{sec:adp}, a parameter update process based on gradient information $\nabla_{\vec{\beta}_i}\hat{v}_{i, {t_k}}$ and $\nabla_{\vec{\beta}_i}K$ is introduced. We propose to use this gradient information to model the ``human action model parameter uncertainty,'' where we assume that the true action model parameter lies in a parameter distribution $\vec \beta_i \sim \mathcal N (\hat {\vec{\beta}}_i,C)$. Here, $\hat {\vec{\beta}}_i$ is the current action model parameter and $C=diag(w^\top abs(\nabla_{\vec{\beta}_i}\hat{v}_{i, {t_k}}+\nabla_{\vec{\beta}_i}K))$ where $w$ is a weighting vector.

To estimate the action completion time, we draw $n$ samples of $\vec \beta_i$ from the distribution $\vec \beta_i \sim \mathcal N (\hat {\vec{\beta}}_i,C)$ and calculate $n$ human completion time for each action according to these sampled parameters. This gives us a distribution of the predicted completion time and allows us to calculate the standard deviation $\sigma_{t_0}$ and $\sigma_{t_h}$ of the sampled completion time. We assume that the human action completion time can be described with a normal distribution, i.e., $t_0\sim \mathcal N (\bar t_0, \sigma_{t_0})$ and $t_h\sim \mathcal N (\bar t_h, \sigma_{t_h})$. Since the 3-sigma bound covers most uncertainty mass, it is acceptable to model $t_0$ and $t_h$ as a random variable taking values in the segment $[\bar t_0 - 3\sigma_{t_0}, \bar t_0 + 3\sigma_{t_0}]$ and $[\bar t_h - 3\sigma_{t_h}, \bar t_h + 3\sigma_{t_h}]$. 

To construct the uncertainty set, we first considered the budget uncertainty \cite{ben2009robust}:
\begin{equation}\label{eq:budget}
\mathcal Z = \{\xi \in \mathbb R^L : -1\leq \xi_l \leq 1,\: l = 1,\dots,L, \: \sum_{l=1}^{L}|\xi_l|\leq \gamma\}, 
\end{equation}
where $\gamma = \sqrt{2ln(1/\epsilon)L}$. Notice that $L$ in this work equals to one plus the number of parallel actions that the human worker can execute, i.e., $L = 1+|C_h|$ where $|\cdot|$ denotes the cardinality of a set. Let $\sigma_0 = 3\sigma_{t_0}$ and $\sigma_h = 3\sigma_{h_0}$, we can rewrite the human action completion time $t_0$ and $t_h$ in~\eqref{eq:TV_opt} as $t_0 = \bar t_0 + \sigma_0\xi_1$ and $t_{h,i} = \bar t_{h,i} + \sigma_h\xi_{i+1}, i = 1,\dots,k$, respectively, where $\xi \in \mathcal Z$. 

\subsection{Robust Task Planning}
With the new representation of the human action completion time, we can formulate the following robust optimization problem \cite{ben2009robust} by introducing additional variables $z , w \in \mathbb R^L$.
\begin{equation}\label{eq:RO_opt}
\begin{aligned}
\min_{x_h,x_r,t,z,w} \quad & t, \\
\textrm{s.t.} \quad & x_r^\top t_r \leq t,   \\
  & \sum_l^{L} |z_l| + \gamma \max_l |w_l| + x_h^\top \bar t_h + \bar t_0\leq t,& \\
  & z_1 + w_1  = \sigma_0,\\
  & z_{i+1} + w_{i+1} = -\sigma_{h,i}x_{h,i},& \;i = 1,\dots,|C_h|, \\
  & x_h + x_r = \textbf{1},\quad x_{h,i}, x_{r,i}\in\{0,1\},& \;i = 1,\dots,k, \\
  & x_{h,\{C-C_h\}}=\textbf{0}, \quad x_{r,\{C-C_r\}}=\textbf{0}.
\end{aligned}
\end{equation}
Notice that this optimization problem is a mixed-integer linear programming problem which can be solved efficiently by commercial solvers, e.g., {\tt\small intlinprog} from {\tt\small MATLAB}, which gives the proposed method a computational advantage. As opposed to search-based task planning method, the optimization formulation allows the solver to approximate the problem as a normal linear programming (LP) problem and only resort to enumerating all possible combinations when the approximation fails. During our simulation experiments, the solver can always solve the problem by solving the approximated LP problems in 0.01 second on average, without the need of enumerating all possible combinations. This shows another advantage of such robust task planning formulation.

\begin{figure}[t]
      \centering
      \includegraphics[width=9cm]{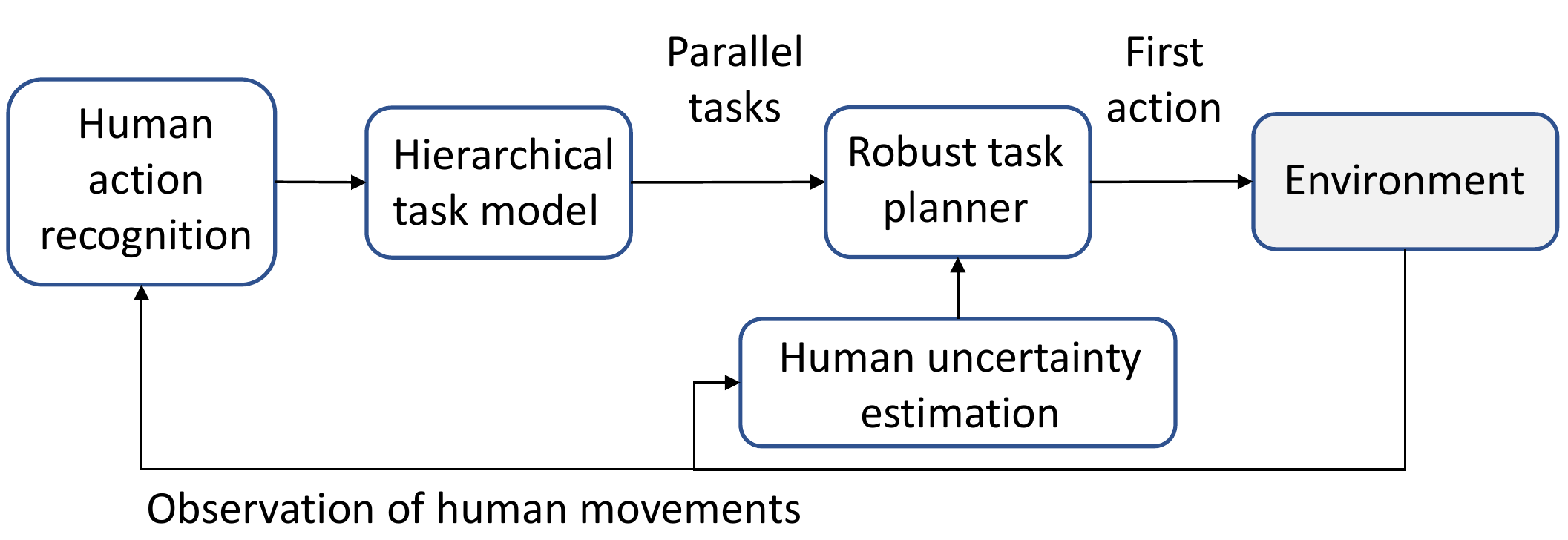}
      \caption{The overall system control design. }
      \label{fig:system}
\end{figure}

The overall system is shown in FIGURE~\ref{fig:system}. During run-time, the robot first collects human motion observations then conducts human action recognition. Given the human's current action, the hierarchical task model finds the parallel actions and send them to the robust task planner. On the other hand, the human uncertainty estimation module, containing the off-line trained human action models, monitors the human uncertainty and updates its models as well as $\sigma_0$ and $\sigma_h$ in the planner. After the plan is generated, the action that has the shortest completion time among all actions assigned to the robot will be executed. This process repeats every time the robot completes its action until there is no remaining action that the robot can execute.

\section{SIMULATION RESULTS}

\begin{figure}[t]
      \centering
      \includegraphics[width=5.5cm]{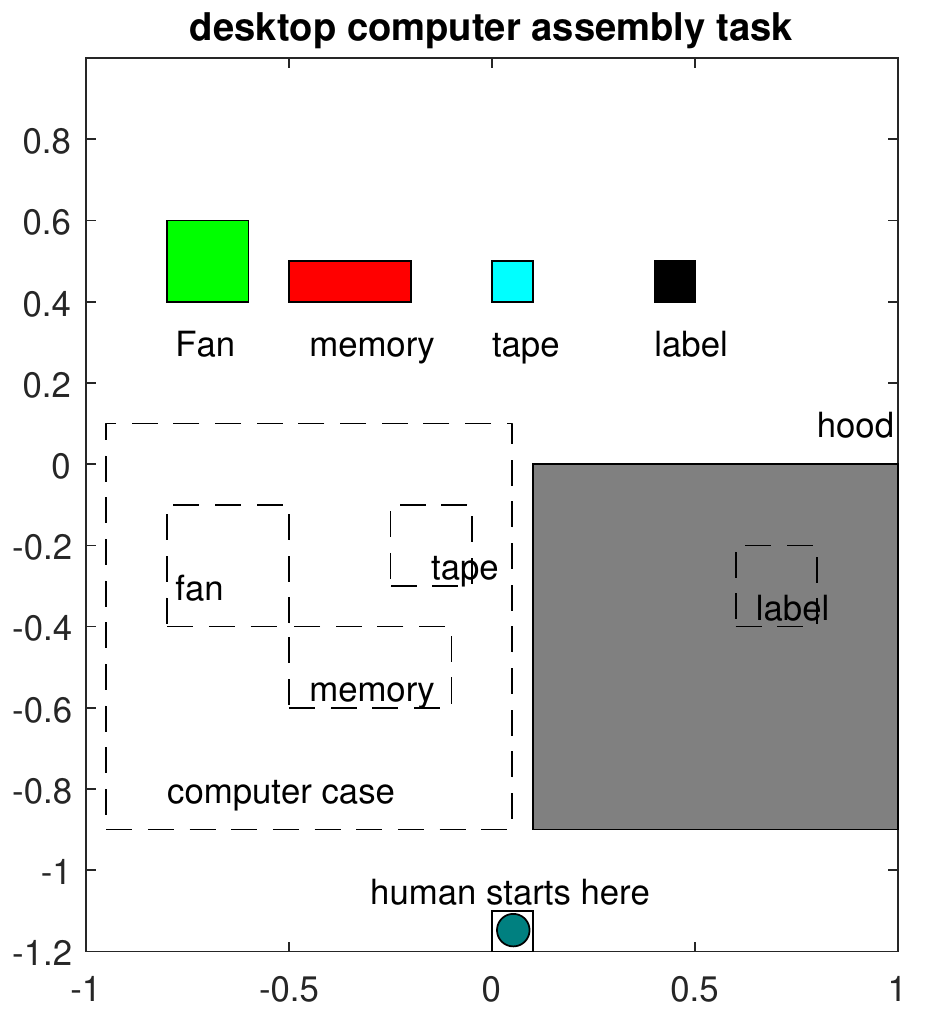}
      \caption{The simulation setup for the computer assembly. The area enclosed by the dashed lines are the designated areas for the objects. The mouse controlled by human is indicated by the green circle.}
      \label{fig:setting}
\end{figure}

The simulation scenario in this work is similar to that of human and robot collaborating for a computer assembly task. We test our algorithm on a simulator built in {\tt\small MATLAB}. Human subjects manipulate the objects using the computer mouse, dragging and releasing the objects with mouse bottom held down and released. Complex actions such as inserting and wrapping in the desktop assembly scenario are simplified by the releasing. The robot actions are expressed by object moving and a sentence shown in the interaction window. As shown in FIGURE~\ref{fig:setting}, five actions are needed to move the fan, the memory, the tape, the label, and the hood to their designated areas. The capability of both the human and the robot is set to the whole action space except that only the human can close the hood. The simulation is conducted in \texttt{Matlab R2020a} on a desktop with 3.2GHz Intel Core i7-8700 CPU. Three case studies with different human worker behaviors are presented in the following sections. Two task planners are considered, the proposed robust task planner and the baseline task planner that treats human action completion time as constants \cite{cheng2021human}. During run-time, the task operation time is shown on the top right and the indicator on top of each object indicates the planner's assignment for the object, where ``h'' indicates that the associated action is assigned to the human worker and ``r'' indicates that the associated action is assigned to the robot. The task completion time for each case is shown in the following table.

\begin{center}
\begin{tabular}{|c|c|c|c| } 
 \hline
 Planner & Case 1 & Case 1 & Case 2 \\ 
 \hline
 Robust task planner & 7.88 [s] & 16.81 [s] &  17.06 [s]\\ 
 Baseline planner & 7.94 [s] & 21.52 [s] &  19.52 [s]\\ 
 \hline
\end{tabular}
\end{center}

\begin{figure*}[t]
 \centering
 \subfloat[Experimental result of the robot with the baseline planner collaborating with an efficient human worker. (Case 1.)\label{fig:mma}]{\begin{tabular}{@{}cccc@{}}
  
   \begin{minipage}{.22\textwidth}
    \includegraphics[width=1\textwidth]{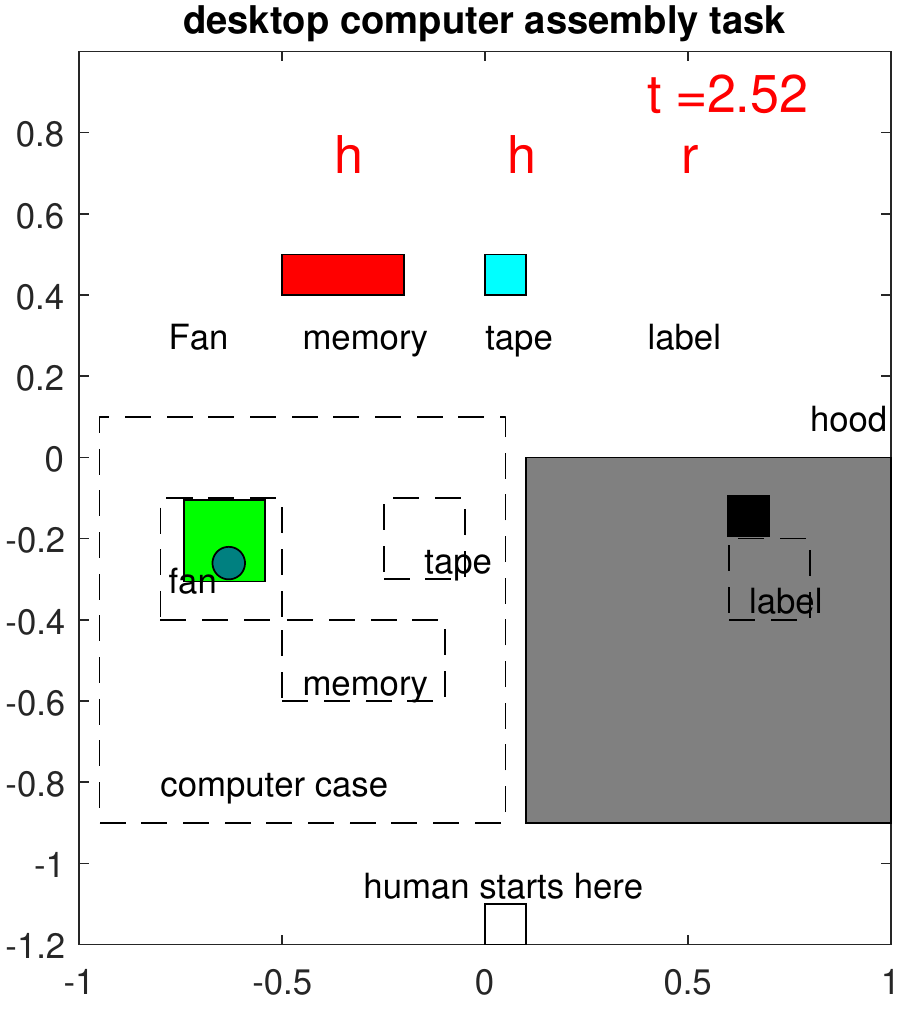}
   
   \end{minipage} &
    \begin{minipage}{.22\textwidth}
    \includegraphics[width=1\textwidth]{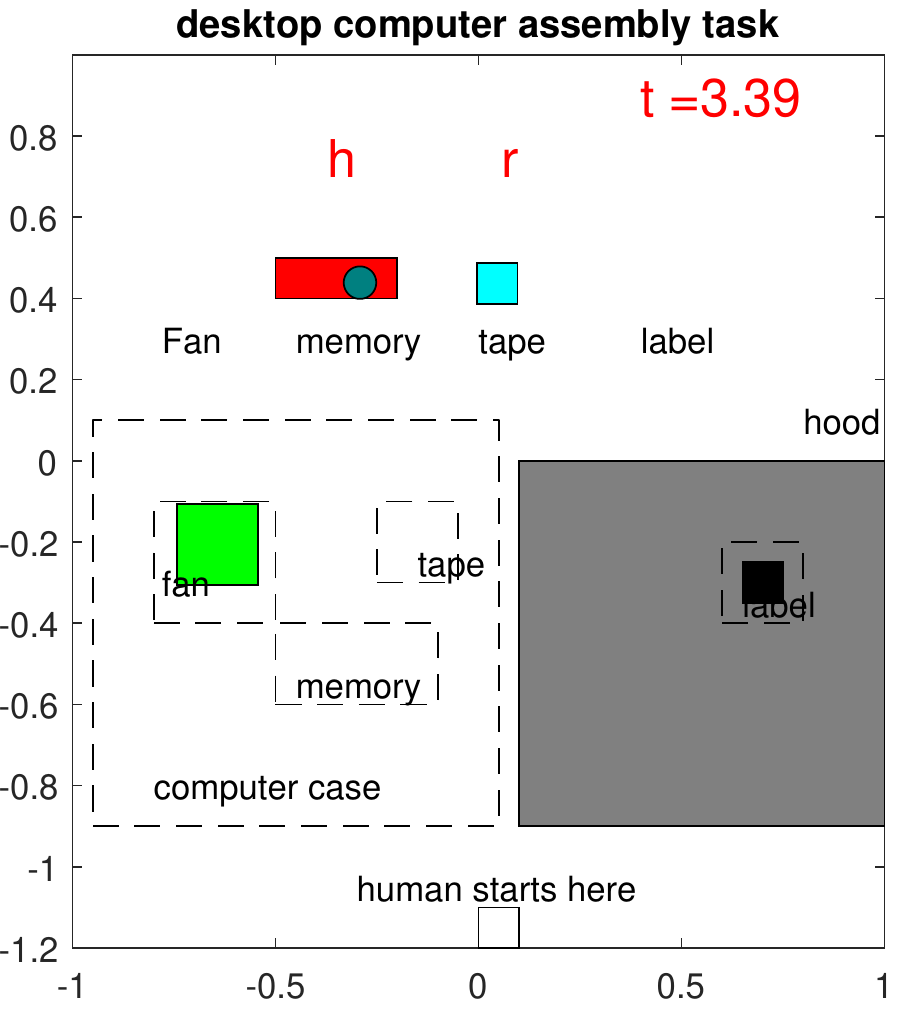}
   
   \end{minipage} &
      \begin{minipage}{.22\textwidth}
    \includegraphics[width=1\textwidth]{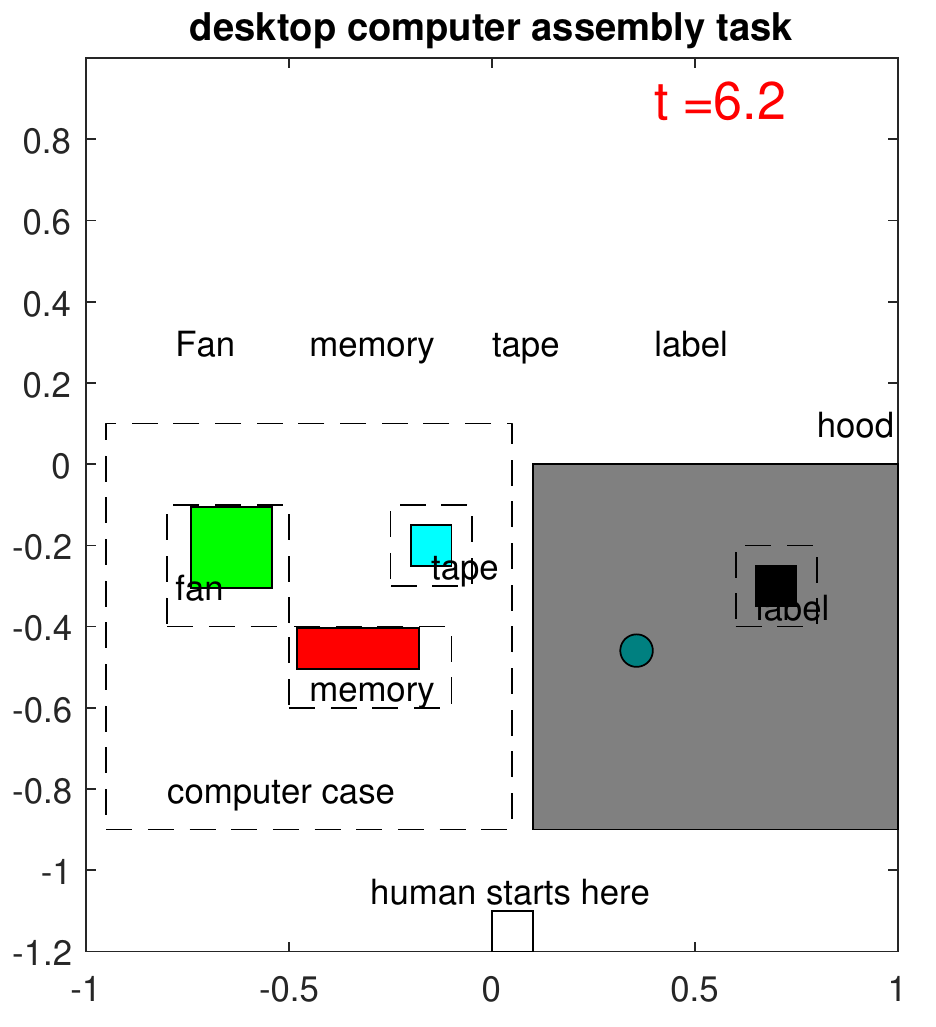}
   
   \end{minipage} &   
      \begin{minipage}{.22\textwidth}
    \includegraphics[width=1\textwidth]{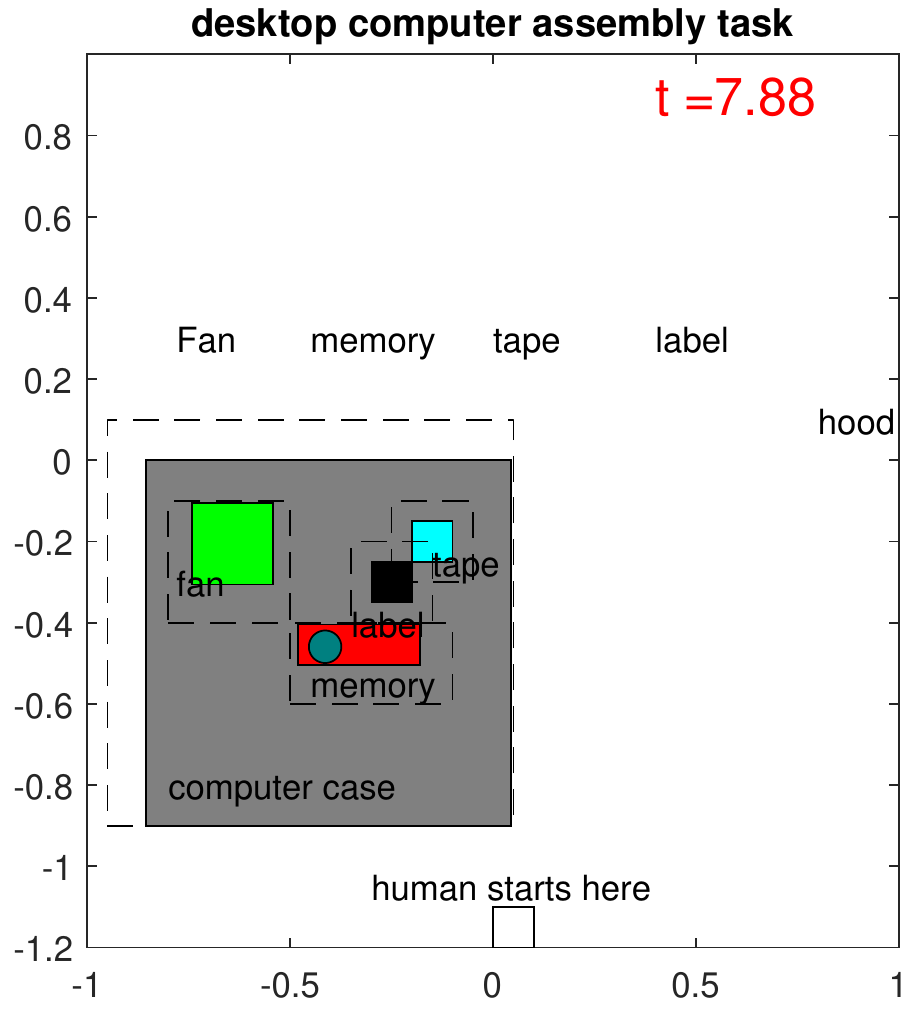}
    
   \end{minipage}\\
  \end{tabular}
  \label{fig:ex_efficient}
  }\\
  
    \subfloat[Experimental result of the robot with the baseline planner (upper row) and the robust task planner (lower row) collaborating with a lazy human worker. (Case 2.)\label{fig:mmc}]{\begin{tabular}{@{}cccc@{}}
  
   \begin{minipage}{.22\textwidth}
    \includegraphics[width=1\textwidth]{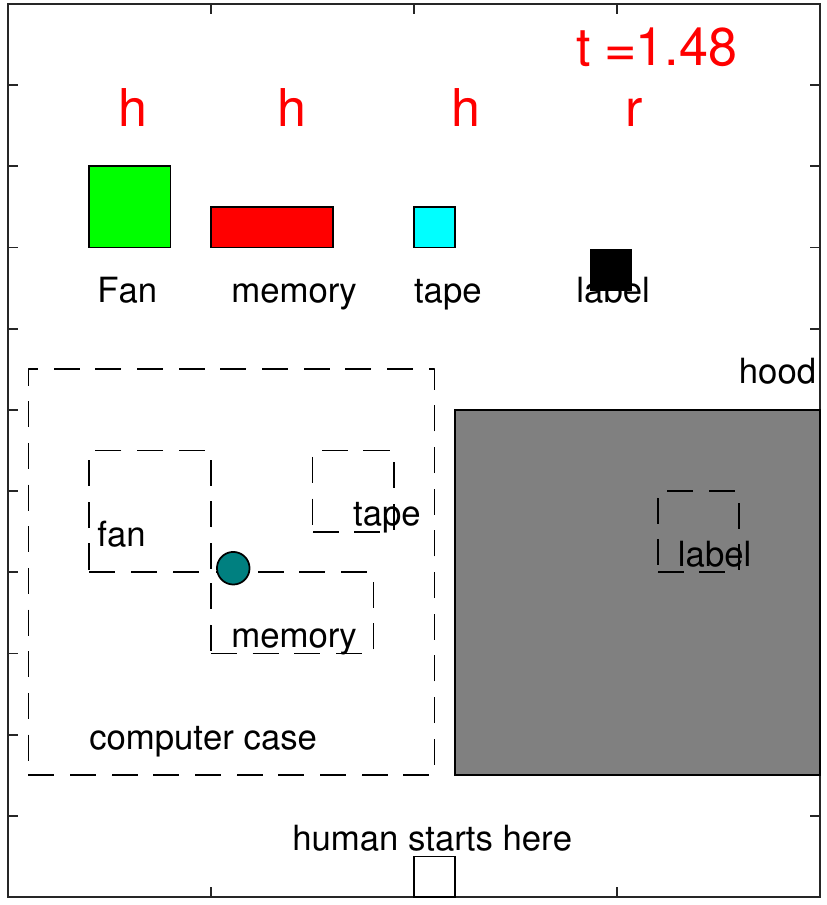}
   
   \end{minipage} &
    \begin{minipage}{.22\textwidth}
    \includegraphics[width=1\textwidth]{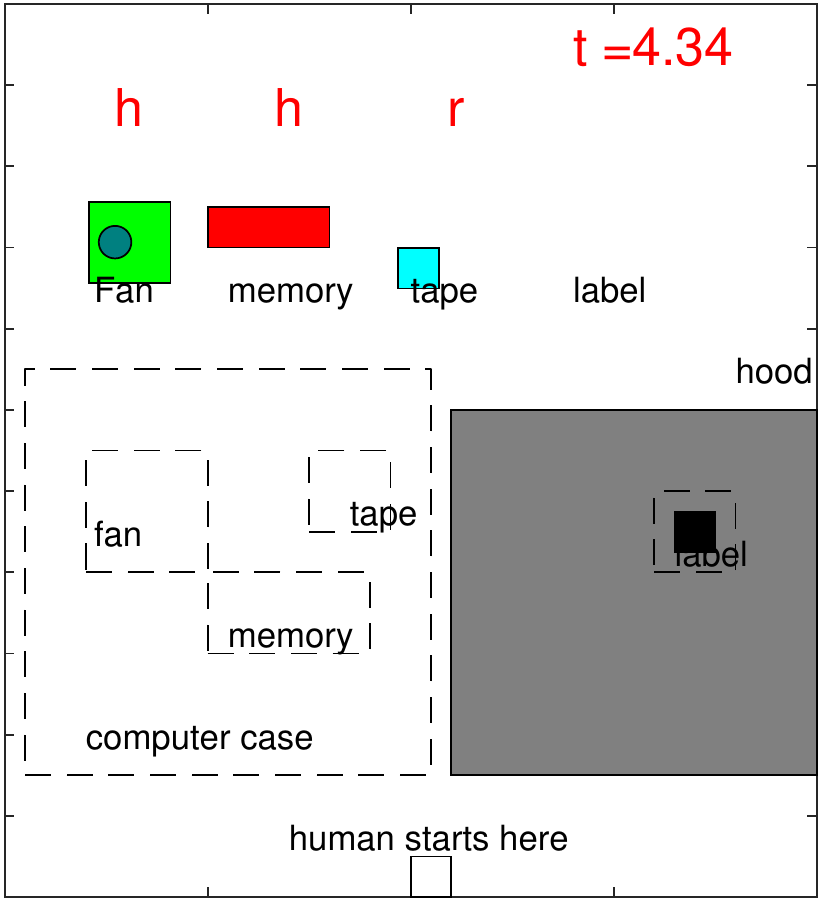}
   
   \end{minipage} &
      \begin{minipage}{.22\textwidth}
    \includegraphics[width=1\textwidth]{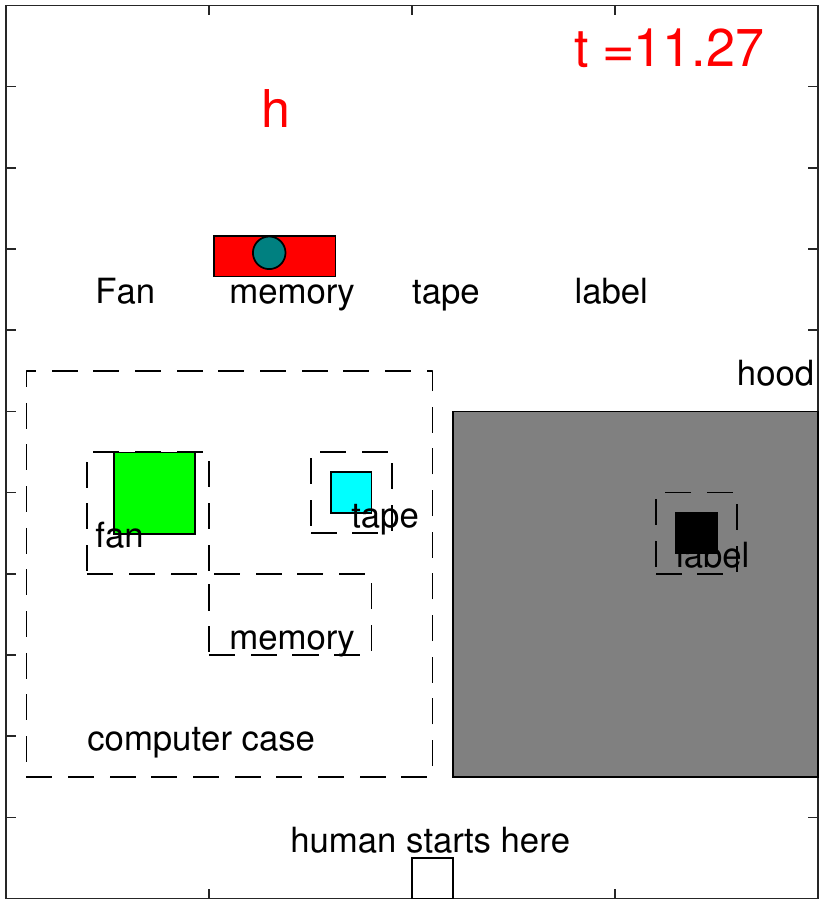}
   
   \end{minipage} &   
      \begin{minipage}{.22\textwidth}
    \includegraphics[width=1\textwidth]{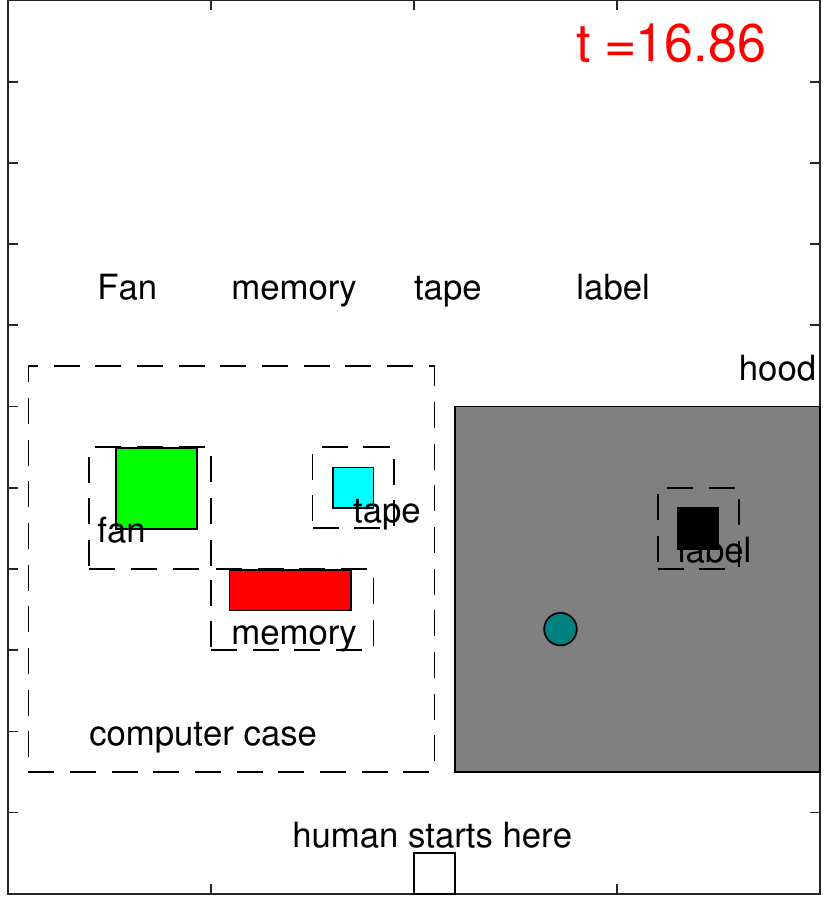}
    
   \end{minipage}\\
   \begin{minipage}{.22\textwidth}
    \includegraphics[width=1\textwidth]{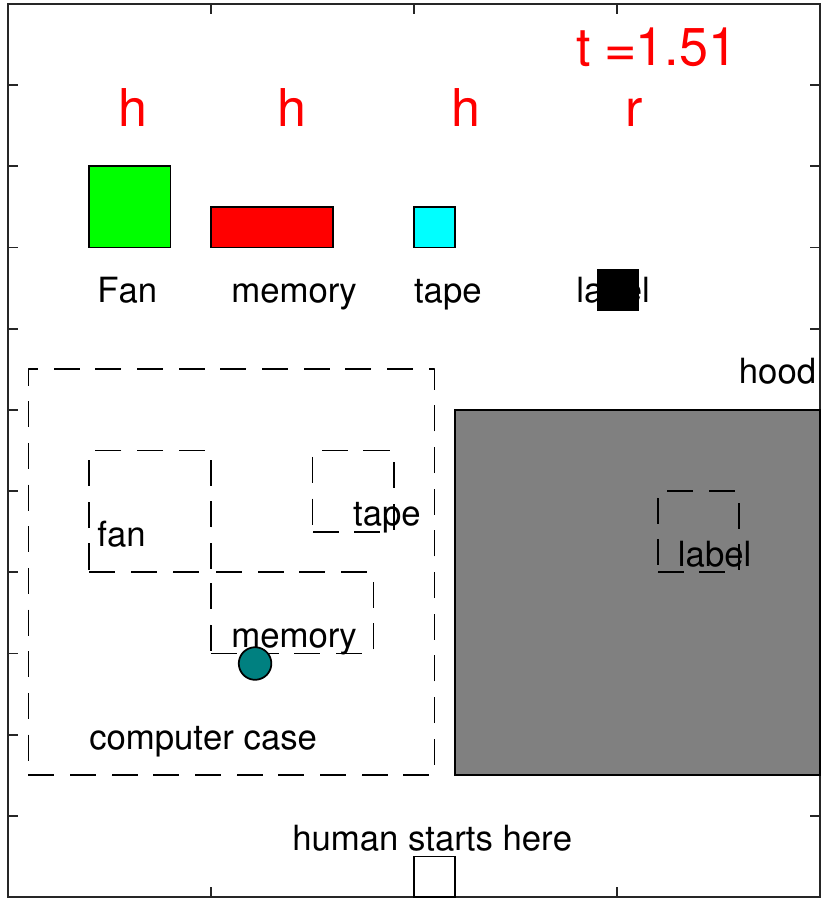}
   
   \end{minipage} &
    \begin{minipage}{.22\textwidth}
    \includegraphics[width=1\textwidth]{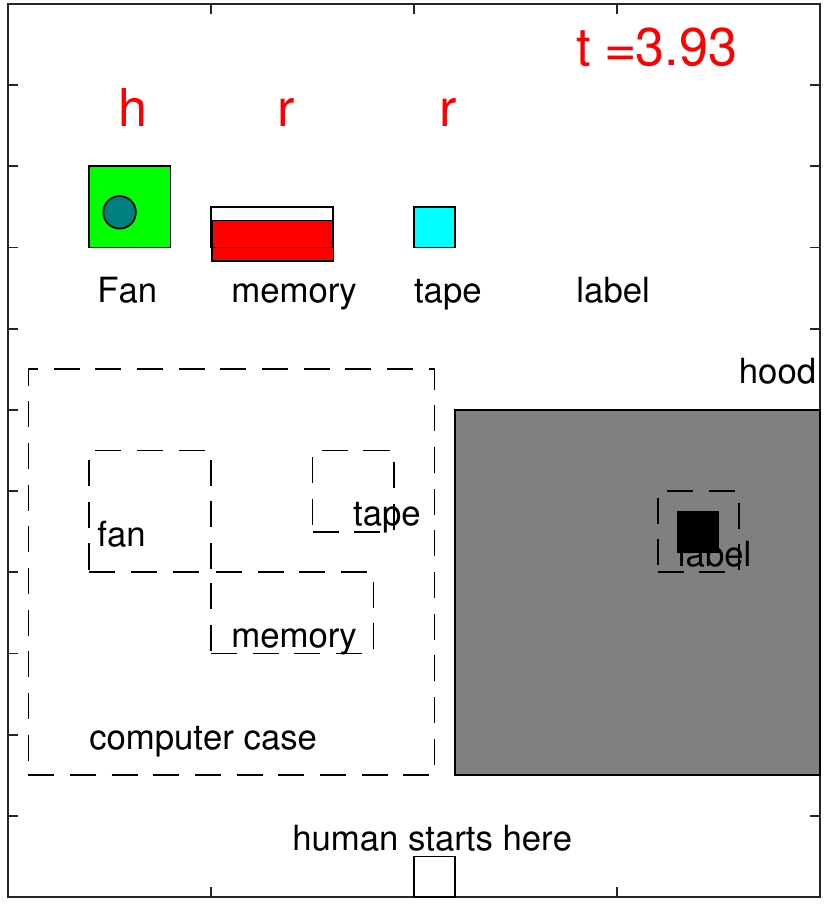}
   
   \end{minipage} &
      \begin{minipage}{.22\textwidth}
    \includegraphics[width=1\textwidth]{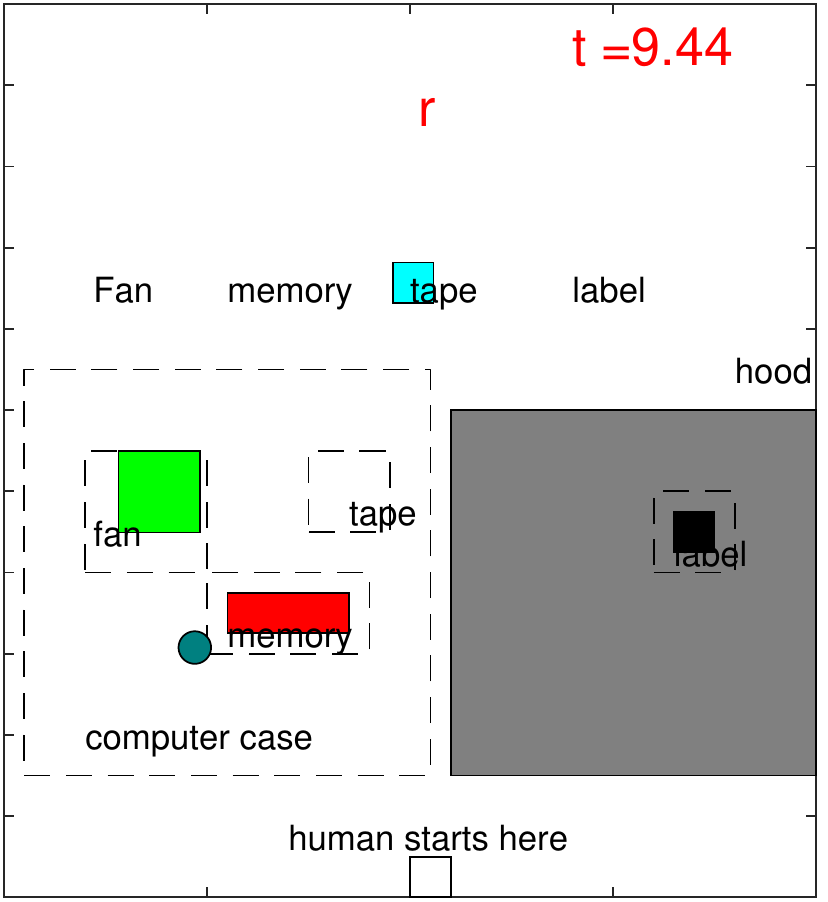}
   
   \end{minipage} &   
      \begin{minipage}{.22\textwidth}
    \includegraphics[width=1\textwidth]{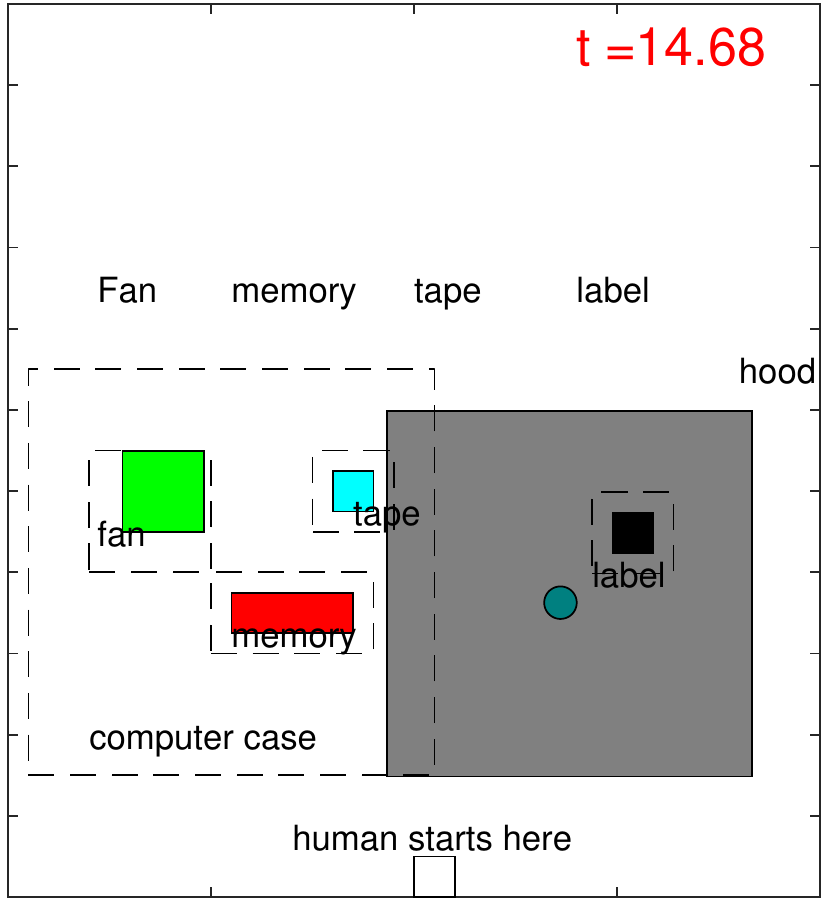}
    
   \end{minipage}\\
  \end{tabular}
  \label{fig:ex_lazy}}\\
  \subfloat[Experimental results of the robot with the robust task planner collaborating with a slacking human worker. (Case 3.)\label{fig:mmc_3}]{\begin{tabular}{@{}cccc@{}}
  
   \begin{minipage}{.22\textwidth}
    \includegraphics[width=1\textwidth]{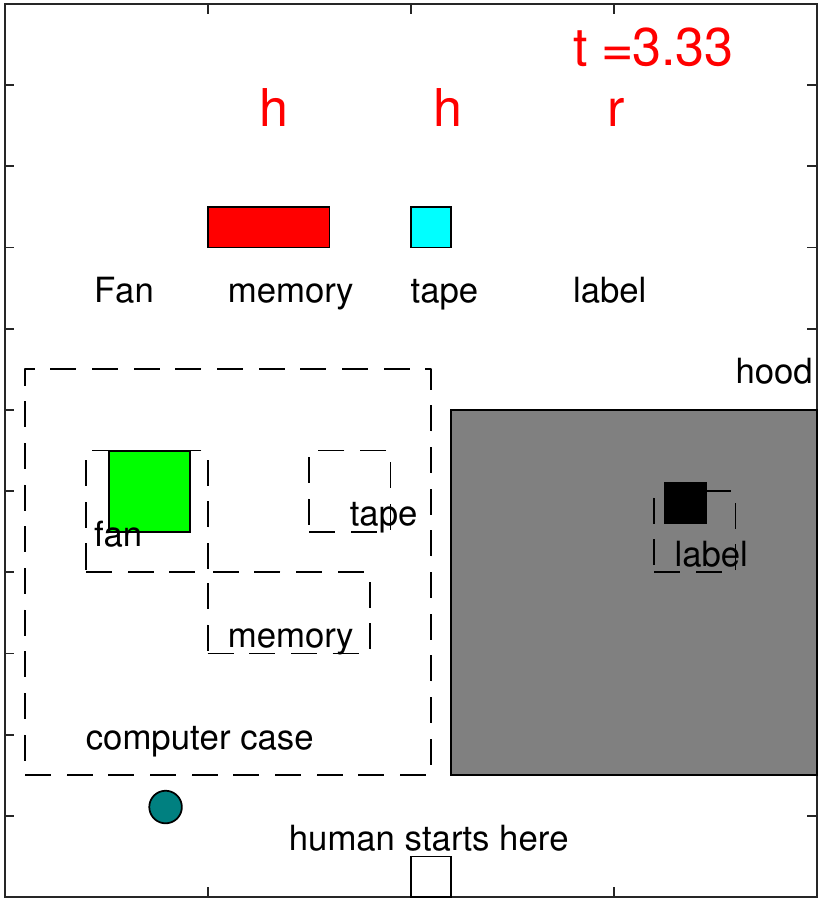}
   
   \end{minipage} &
    \begin{minipage}{.22\textwidth}
    \includegraphics[width=1\textwidth]{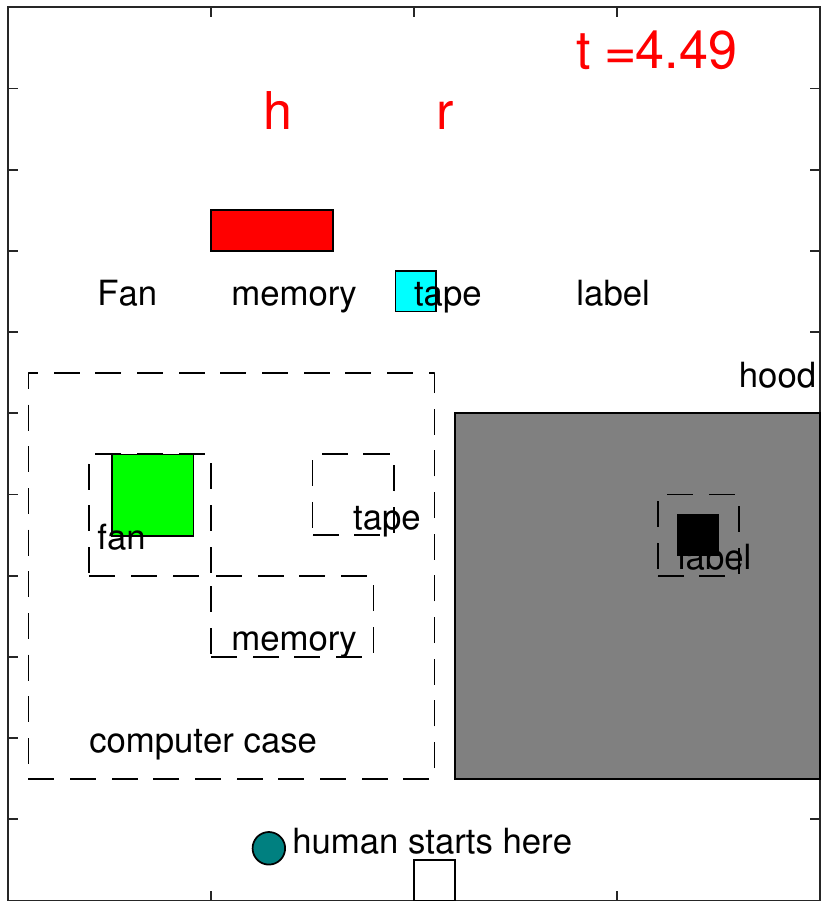}
   
   \end{minipage} &
      \begin{minipage}{.22\textwidth}
    \includegraphics[width=1\textwidth]{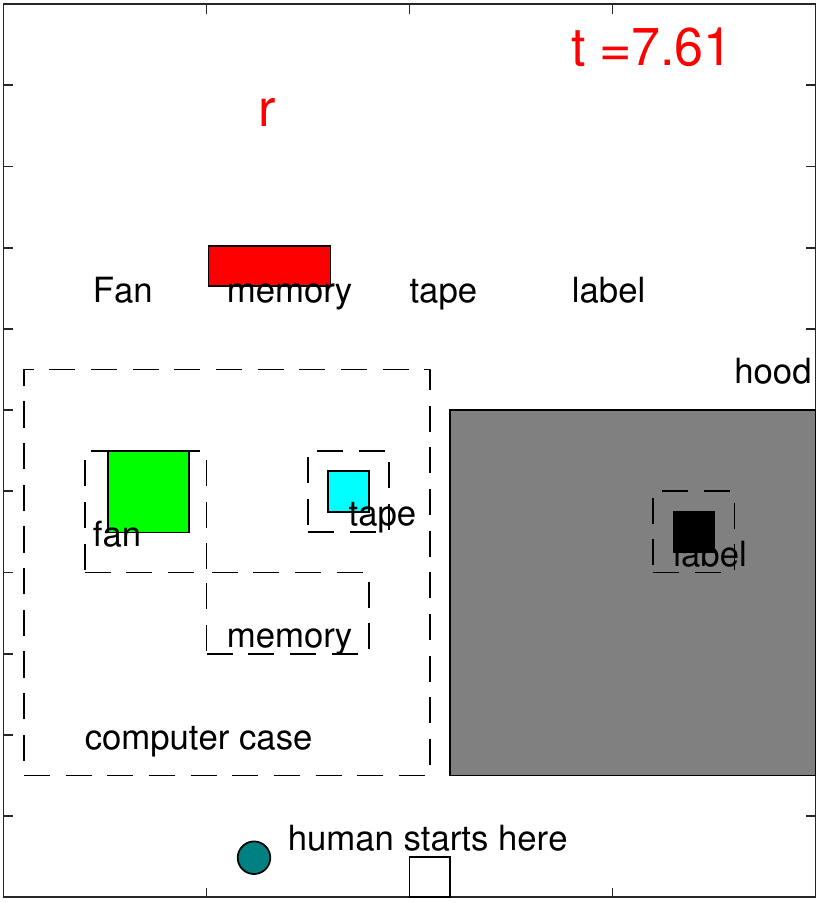}
   
   \end{minipage} &   
      \begin{minipage}{.22\textwidth}
    \includegraphics[width=1\textwidth]{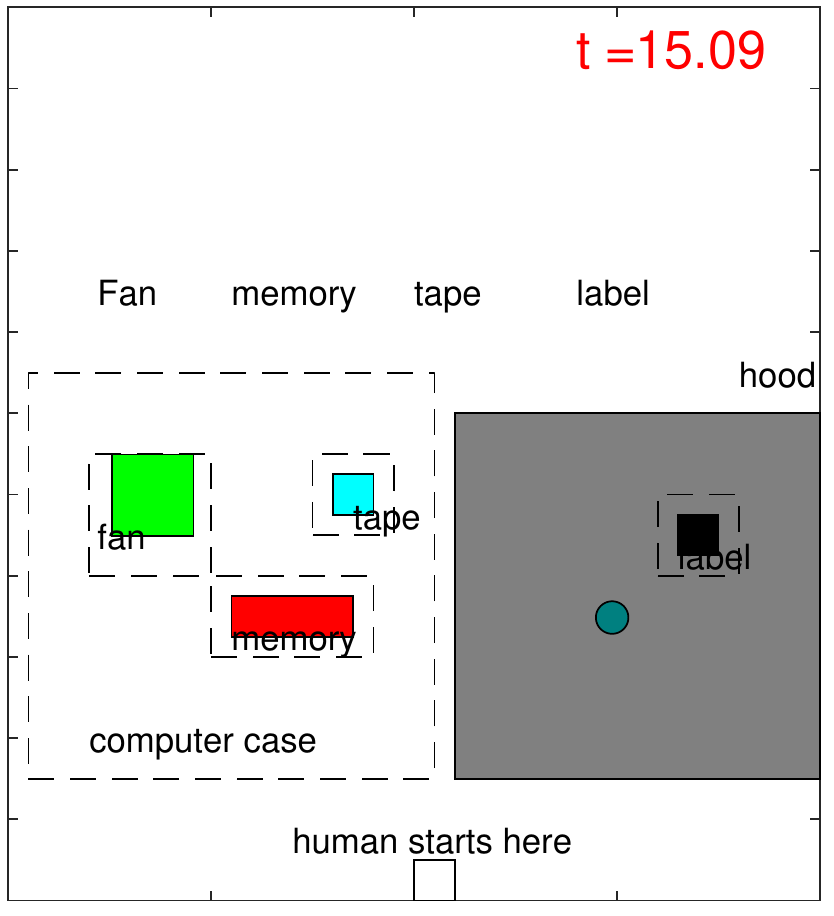}
    
   \end{minipage}\\
  \end{tabular}
  \label{fig:ex_slack}}\\
  
  \caption{Experimental results of a robot collaborating with different human workers.}
  \label{fig:ex}
\end{figure*}

\noindent\textbf{Case 1: collaborating with an efficient human worker.}
The word ``efficient'' means that the worker performs each action using the amount of time similar to the prior knowledge that the planner was given. As shown in FIGURE~\ref{fig:ex_efficient}, once the planner detects the human's current action, the planning is conducted using the given human completion time. The resulting plan lets the human worker to place the memory and the tape, while assigning the robot to place the label. After the robot finishes placing the label, the planner replans and realizes that the robot should move the tape since the human just starts to move the memory. At the end, human worker closes the hood and finishes the task. In this case we can see that the baseline planner collaborated with the worker efficiently. A similar behavior can be seen when the robot runs the proposed robust task planner. 

\noindent\textbf{Case 2: collaborating with a lazy human worker.}
The word ``lazy'' means that the worker performs each action using 2x to 3x the amount of time comparing to the prior knowledge that the planner was given. The upper row in FIGURE~\ref{fig:ex_lazy} shows the result of robot running with the baseline planner. Since the completion time is treated as constants, the planner continues to assign the memory to the human because it believes the human worker can complete that task quicker than the robot. This causes the robot to idle while the human slowly completes the current action, then completes the the memory placement. On the other hand, our proposed method detects the deviation of the human movements form the model, therefore, it assigns both the tape and the memory to the robot once the planner finishes replanning, as shown in the lower row in FIGURE~\ref{fig:ex_lazy}. This shows that our planner is able to account for human-induced uncertainties and as a result, reduce the damage of these uncertainties in terms of task completion time.

\noindent\textbf{Case 3: collaborating with a slacking human worker.}
The word ``slacking'' means that the worker stops doing the task for some period of time in the middle. The worker can either be an efficient worker or a lazy worker. As shown FIGURE~\ref{fig:ex_slack}, the mouse idles around the bottom area of the interaction window for sometime ($t=4\sim 13$s). When using the baseline planner, similar to that happens in case 2, the planner continues to assign the memory to the human worker after the fan, tape, and label placements are completed. With the proposed planner, the memory placement will be completed by the robot so that the time for the human worker to execute this action is saved. FIGURE~\ref{fig:ex_slack} shows the result where the human worker works ``efficiently'' most of the time but slack off sometimes during the task. Since the human moves fast (completes the fan placement in $3$ seconds), the planner originally assigns memory placement to the human when the robot is placing the tape. However, the planner later on assigns this to the robot because the human worker is idling and causes larger human uncertainties. This demonstrates the ability of the proposed task planner to handle abnormal human behaviors.

In summary, the proposed robust task planner is able to take in the human uncertainty estimation and replan accordingly, resulting in shorter task completion time when human action deviates from the prior knowledge.

\section{CONCLUSION}

This paper presented a robust task planner for assembly lines with human robot interaction. Based on the adaptation process of the off-line-learned human action models, we sampled the model parameters and estimated the distribution of the human action completion time to quantify the human-induced uncertainty. The robust task planner took in this information, alongside with the hierarchical task model and solved the task planning problem as a robust optimization problem. Simulation results showed that the proposed robust task planner was able to handle the human uncertainty and replan accordingly, resulting in a shorter task completion time comparing to that of the baseline planner when human action deviated from the prior knowledge. Future works include incorporating a real robot and a human model into the simulation environment and conducting real-world experiments. 

\clearpage


\bibliographystyle{asmems4}

\begin{acknowledgment}
This work was supported by the National Science Foundation under Grant No.1734109. Any opinion, finding, and conclusion expressed in this paper are those of the authors and do not necessarily reflect those of the National Science Foundation. 
\end{acknowledgment}

%
% ...

\bibliography{ISFA2022}

\end{document}